\def\year{2021}\relax

\documentclass[letterpaper]{article} 
\usepackage{aaai21}  
\usepackage{times}  
\usepackage{helvet} 
\usepackage{courier}  
\usepackage[hyphens]{url}  
\usepackage{graphicx} 
\usepackage{natbib}  
\usepackage{caption} 
\usepackage{arydshln}
\usepackage[switch]{lineno}
\title{Towards Faithfulness in Open Domain Table-to-text Generation 
\\from an Entity-centric View}
\author {
        Tianyu Liu \textsuperscript{\rm 1} \thanks{The first two authors contribute equally to this work.}, Xin Zheng \textsuperscript{\rm 2 3} \footnotemark[1], Baobao Chang \textsuperscript{\rm 1 4}, Zhifang Sui \textsuperscript{\rm 1 4}\\
}
\affiliations {
    \textsuperscript{\rm 1} Ministry of Education (MOE) Key Laboratory of Computational Linguistics, School of EECS, Peking University \\
    \textsuperscript{\rm 2} Chinese Information Processing Laboratory, Institute of Software, Chinese Academy of Sciences, Beijing, China \\
    \textsuperscript{\rm 3} University of Chinese Academy of Sciences, Beijing, China
    \textsuperscript{\rm 4} Pengcheng Laboratory, Shenzhen, China \\
    
    tianyu0421@pku.edu.cn, zheng\_xin@bupt.edu.cn, chbb@pku.edu.cn, szf@pku.edu.cn
}

\begin{document}
\maketitle
\begin{abstract}
In open domain table-to-text generation, we notice that the unfaithful generation usually contains hallucinated content which can not be aligned to any input table record.
We thus try to evaluate the generation faithfulness with two entity-centric metrics: table record coverage and the ratio of hallucinated entities in text, 
both of which are shown to have strong agreement with human judgements.
Then based on these metrics, 
we quantitatively analyze the correlation between training data quality and generation fidelity which indicates the potential usage of entity information in faithful generation.
Motivated by these findings, we propose two methods for faithful generation: 1) augmented training by incorporating the auxiliary entity information, including both an augmented plan-based model and an unsupervised model and 2) training instance selection based on faithfulness ranking. We show these approaches 
improve generation fidelity in both full dataset setting and few shot learning settings by both automatic and human evaluations.
\end{abstract}

\section{Introduction}
\label{intro}

The difficulty of faithful table-to-text generation originates from the divergence of source tables and reference text \cite{perez2018bootstrapping} in the training stage, especially for open domain datasets without human curation, e.g. \textsc{Wikiperson} \cite{wang2018describing}.
In \textsc{Wikiperson}, we observe that the unfaithful generation, as well as some reference text in the training data, often contains hallucinated entities which can not be aligned with any input table record.
This motivates us to evaluate the faithfulness of a text to a given table by two entity-centric metrics: 
the statistics of table record coverage
and hallucinated entities in both training and evaluation stages.
The metrics are computed based on the name entities recognition (NER) in text with off-the-shelf tools, and their alignment with table records.
We find the proposed metrics have high correlation with human judgements (Sec \ref{sec:entity_centric_evaluation}).  

Then we quantitatively study how training data quality affects the generation fidelity (Sec \ref{sec:quantitative_analysis}) based on the proposed metrics.
Specifically we show that filtering uncovered records in source table would increase generation coverage\footnote{Measuring generation coverage can be useful in some specific scenarios. The generator would be expected to cover all records (e.g. \textsc{WebNLG} \cite{colin2016webnlg}) or only summarize salient records (e.g. \textsc{WikiBio} \cite{lebret2016neural}).}, and truncating the sentences prone to hallucination in reference text would accordingly reduce generation hallucination. 
However, we also observe a trade-off between generation faithfulness and coverage, 
e.g. truncating reference text would possibly lower the generation coverage to source table.
We thus seek for an alternative method to reduce the hallucinations in the training stage while maintaining high coverage.


To this end, we firstly propose plan-based generation with pseudo parallel training corpus, which we call the augmented plan-based method (Sec \ref{sec:augmented_plan}). Some prior work \cite{moryossef2019step,ferreira2019neural} have shown that the two-stage plan-based generation could lead to a more accuracy and controllable generation in data-to-text generation. However, most previous work focus on high quality datasets without much content mismatch between the input table and the reference text (e.g. \textsc{WebNLG} and \textsc{E2E} \cite{dusek2019e2e}). However, in the open domain setting, we have to face noisy training data \footnote{The main reason for data noise is the lack of human intervention in data collection. It could be of high cost to monitor the faithfulness of multi-sentence reference in the open domain.}, i.e. hallucinations in reference text. Thus we exploit serialized plans which are formulated as text sequences, in this way, we can easily use any established sequence-to-sequence model as neural planner rather than a dedicated planning module \cite{moryossef2019step,moryossef2019improving}. Our plans model the number of sentences to be generated and the order of given table records in each sentence.
In the plan-to-text phase, we create a pseudo parallel training corpus which incorporates the auxiliary uncovered entities in the reference text into an augmented plan, and then the model generates text according to the augmented plan. The augmented plans effectively reduce the hallucinations in the pseudo training corpus and correspondingly encourage the generator to stick to the designated plans.
The experiments show that the proposed augmented plan based methods not only reduce about 60\% hallucinations compared with the counterparts trained in the end-to-end fashion in full dataset setting but also greatly enhance the model performance in few shot learning. We also achieve the state-of-the-art performance on \textsc{Wikiperson} dataset and show that the proposed methods work well on both pre-trained models and (non pre-trained) Transformer models.

We also propose a training instance selection method based on proposed hallucination evaluation metrics and related faithfulness ranking (Sec \ref{sec:instance-select}). We show that with only 5\% top ranking training instances,
pretrained model BART \cite{lewis2019bart} outperforms its variant trained on full dataset by about 7 PARENT \cite{dhingra2019handling} points. This also implies the effectiveness of the proposed entity-based metrics for faithfulness evaluation.

\section{Faithfulness in Table-to-text}
\label{metric}
In this section, we introduce two evaluation metrics called table record coverage $P_{cover}$ and the ratio of hallucination $R_{hallu}$ based on the alignment between source table records and recognized entities in the text. Through human annotations, we verified that the proposed metrics have high correlation with human judgement. Finally, we practice quantitative analyses to show how the proposed two metrics correlate with generation fidelity and gain insights on how to increase table coverage or reduce hallucinations in open domain table-to-text generation.
\subsection{Entity-centric Evaluation}\label{sec:entity_centric_evaluation}
A structured table $T$ can be expressed by a set of records $\{r_k\}_{k=1}^K$, where each record is an \emph{(attribute, value)} pair. There is also a reference description $R$ available for each table.
The task aims at generate a text $G$ which describes the records in $T$ fluently and faithfully.
Due to the lack of human intervention, the reference $R$ may contain hallucinations and have low table records coverage in the open domain datasets.

\subsubsection{Entity recognition and alignment}
With off-the-shelf NER tools\footnote{We use Stanza \cite{qi2020stanza} toolkit throughout our experiments. \url{https://stanfordnlp.github.io/stanza/}}, we could recognize entities like \emph{person}, \emph{location}, \emph{date}, \emph{event} in text, which characterize the expressed facts. 
Suppose we have recognized a group of entities $\{e_i\}_{i=1}^{M}$ from a specific text $G$ ($G$ can be training reference $R$, which aims at evaluating training data quality or predicted table description $D$, which aims at assessing model performance on test sets.). We then heuristically decide whether a table record $r_k$ is \textbf{covered} by $G$.

Our heuristic matching rules include: 1) Exact Match: a text span in $G$ can exactly match $r_k$'s \emph{value}. 2) Filtered Sub-sequence Match: For non-numeric entity $e_i$ with \emph{PERSON, NORP, FAC, ORG, GPE, LOC, PRODUCT, EVENT or WORK\_OF\_ ART} labels, after filtering the stop words in $e_i$ and $r_k$'s value, $r_k$ matches $e_i$ if and only if $e_i^{filter}$ is a sub-sequence of $r_k^{filter}$.


In our benchmark dataset \textsc{Wikiperson}, 99.3\% of table records in the training set could be aligned with corresponding reference text with the above mentioned heuristic matching strategies. As suggested by training set, the test-time generation could cover most table records without too much pain when using a powerful generator, however reducing the hallucinations in the generated text is non-trivial.

\subsubsection{Entity-centric metrics}
After recognizing $\{e_i\}_{i=1}^{M}$ in text $G$, we use the above mentioned heuristic strategies to establish an alignment between $\{e_i\}_{i=1}^{M}$ and table records $\{r_k\}_{k=1}^K$.
We consider the unaligned entities in $\{e_i\}_{i=1}^{M}$ as hallucinated entities in $G$. For a pair of table and text $(T^j, G^j)$, we define $n^j_{hallu}, p^j_{cover}, l^j$ as the number of hallucinated entities in $G^j$, the ratio of covered records in $T^j$ and the length of $G^j$ respectively.
For a set of $N$ (table, text) pairs $\{(T^j, G^j)\}_{j=1}^N$, 
the corpus level table coverage $P_{cover}$ and the hallucinated ratio $R_{hallu}$ are shown as follows:

$P_{cover}= \sum_j p_{cover}^j / N$;
$R_{hallu}= \sum_j n_{hallu}^j / (N*L)$

\noindent in which $L$=$\sum_j l^j/N$ is the average length of text $\{G^j\}_{j=1}^N$.

\subsubsection{Correlation with human judgement}
To assess the correlation of $P_{cover}$ and $R_{hallu}$ with human judgement, we sample 400 \emph{(table,text)} pairs, of which 100 pairs from training set, 300 pairs from the 3 model outputs (in Table \ref{tab:benchmark}) while testing (100 pairs each). 
We hire three well-educated annotators to decide 1) for each table record whether it is covered by corresponding text and 2) for each recognized hallucinated entity in text whether it is really a hallucination or not.
The Fleiss' kappa between annotators is 0.71, which indicates a `substantial agreement' \cite{landis1977measurement}.
We then use majority vote to decide the final annotations. 
Our coverage and hallucination decisions achieve 95.2\% and 93.7\% accuracy respectively according to human annotations.

\subsubsection{Limitations}
Note that we have shown that proposed metrics have high \emph{precision} in identifying hallucinations. However it is not enough to merely consider recognized entities in text, in other words, the proposed $R_{hallu}$ may not have a high \emph{recall} on recognizing other forms of hallucinations (e.g. verbal hallucination). We thus ask the human judges to mark any hallucinated word in the generated text while comparing different systems.

\begin{table}[t]
\small
\setlength{\tabcolsep}{1mm}{
\begin{tabular}{l|ccccc}
\hline
Model & BLEU & PARENT & $P_{cover}(\uparrow)$ & $R_{hallu}$($\downarrow$) & LEN \\ \hline
PG-Net & 23.56 & 50.14 & 88.63\% & \textbf{0.091} & 59.1  \\
Transformer & 24.63 & 51.86 & 89.74\% & 0.093 & 63.4  \\
BART  & \textbf{31.16} & \textbf{52.54} & \textbf{98.31\%} & 0.099 & 81.6 \\ \hline
\end{tabular}}
\caption{Model performances on the \textsc{Wikiperson} dataset in terms of BLEU-4 \cite{papineni2002bleu}, PARENT and the entity-centric statistics (Sec \ref{sec:entity_centric_evaluation}). } 

\label{tab:benchmark}
\end{table}

\subsection{Quantitative Analysis}\label{sec:quantitative_analysis}
Firstly we benchmark Pointer-Generator (PG-Net) \cite{see2017get}, Transformer \cite{vaswani2017attention} and BART models on \textsc{Wikiperson} dataset in Table \ref{tab:benchmark}.
We then analyze how $P_{cover}$ and $R_{hallu}$ (Sec \ref{sec:entity_centric_evaluation}) in training data affect the generation fidelity.
Note that we do not change the test and dev sets in all following experiments.

\subsubsection{Filtering uncovered table records}
To better investigate how $P_{cover}$ in training set affects generation coverage, we only keep \emph{the first sentence}\footnote{As the original training dataset has 99.3\% table record coverage (Table \ref{tab:pcover}a), `one sentence' setting allows a wider range  (67.7\%-100\% in Table \ref{tab:pcover}a) in controlling $P_{cover}$ by $\lambda$. }
in reference text and drop other sentences while training.
Then we filter the uncovered table records by setting a threshold $\lambda$, e.g. when $\lambda$=$0.75$, we randomly filter 75\% of uncovered records in training set. 

\begin{table}[t!]
\small
\centering

\setlength{\tabcolsep}{1.3mm}{
\begin{tabular}{lccccc}
\hline
 & $\lambda$ = 0 & $\lambda$ = 0.25 & $\lambda$ = 0.5 & $\lambda$ = 0.75 & $\lambda$ = 1.0 \\ \hline
\multicolumn{6}{c}{Table Record Coverage (\%) $P_{cover}(\uparrow)$}\\
Train & 67.7 & 72.5 & \textbf{78.7} & \textbf{87.3} & \textbf{100.0} \\
PG-Net & 69.6 & 71.4 & 73.6 & 77.4 & 85.7 \\
Transformer & 71.2 & 72.8 & 75.2 & 79.0 & 86.3\\
BART & \textbf{75.4} & \textbf{77.2} & 78.1 & 82.6 & 91.9 \\ \hline
\multicolumn{6}{c}{Ratio of hallucinated entities $R_{hallu}(\downarrow)$}\\
Train & 0.060 & 0.060 & 0.060 & 0.060 & 0.060 \\
PG-Net & 0.049 & 0.051 & 0.052 & 0.054 & 0.057\\
Transformer & 0.045 & 0.046 & \textbf{0.047} & \textbf{0.049} & \textbf{0.054}\\
BART & \textbf{0.043} & \textbf{0.045} & 0.048 & 0.053 & 0.063 \\ \hline
\multicolumn{6}{c}{Avg. length of (tokenized) text}\\
Train & 27.7 & 27.7 & 27.7 & 27.7 & 27.7 \\
PG-Net & 25.3 & 25.8 & 26.0 & 27.2 & 28.6\\
Transformer & 25.7 & 26.1 & 26.3 & 27.5 & 29.2\\
BART & 25.5 & 26.0 & 26.6 & 28.2 & 34.7\\ \hline
    \end{tabular}}
\\(a) Filtering uncovered table records 
\setlength{\tabcolsep}{1.3mm}{
\begin{tabular}{lcccc}\hline
 & No Trunc & $N_{keep}$=3 & $N_{keep}$=2 & $N_{keep}$=1 \\\hline
\multicolumn{5}{c}{Table Record Coverage (\%) $P_{cover}(\uparrow)$}\\
Train & \textbf{99.3} & 91.4 & 82.8 & 67.7\\
PG-Net & 88.6 & 85.7 & 78.4 & 69.6 \\
Transformer & 89.7 & 86.9 & 81.4 & 71.2\\
BART & 98.3 & \textbf{96.9} & \textbf{89.6} & \textbf{75.4}\\\hline
\multicolumn{5}{c}{Ratio of hallucinated entities $R_{hallu}(\downarrow)$}\\
Train & 0.096 & 0.086 & 0.077 & 0.060\\
PG-Net & \textbf{0.091} & \textbf{0.078} & \textbf{0.061} & 0.049\\
Transformer & 0.093 & 0.080 & 0.064 & 0.045\\
BART & 0.099 & 0.083 & 0.067 & \textbf{0.043}\\\hline
\multicolumn{5}{c}{Avg. length of (tokenized) text}\\
Train & 88.3 & 62.7 & 46.1 & 27.7\\
PG-Net & 59.1 & 44.7 & 32.3 & 25.3\\
Transformer & 63.4 & 48.2 & 36.2 & 25.7\\
BART & 81.6 & 58.6 & 40.9 & 25.5\\ \hline
    \end{tabular}}
\\(b) Truncating reference text
    \caption{The statistics of training data and model outputs on test set in two different settings (Sec \ref{sec:quantitative_analysis}).}
    \label{tab:pcover}

\end{table}

\begin{figure*}[t!]
    \centering
    \includegraphics[width=0.85\linewidth]{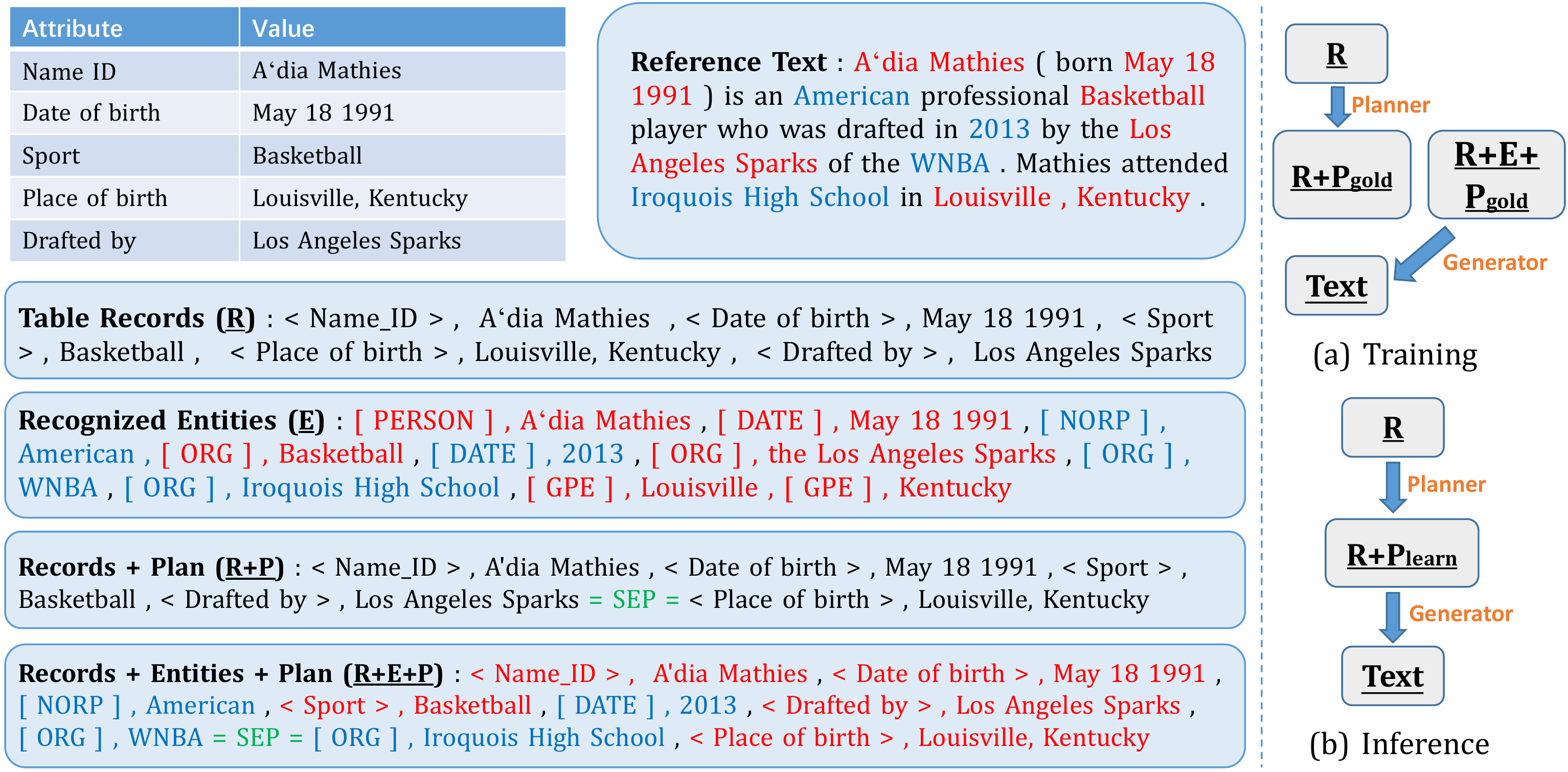}
    \caption{For the left figure, we show different forms of model inputs (Table \ref{tab:plan_results}) for the text generator in the two-phase generation. We mark the covered table records, the recognized hallucinated entities and the sentence delimiter in red, blue and green respectively. For the right figure, we show the training and inference procedures in the augmented plan-based generation (ID=8 in Table \ref{tab:plan_results}).}
    \label{fig:plan_based_generation}

\end{figure*}

\subsubsection{Truncating reference text}
On average, each reference text in \textsc{Wikiperson} training set contains 4.33 sentences. 
We observe that the sentences near the end are prone to hallucination, so we try to reduce $R_{hallu}$ in the training set by truncating the reference text. We let $N_{keep}$ to denote the max number of sentences we keep in each reference text and then drop all the following sentences.

\subsubsection{Insights from quantitative analysis}\label{sec:insights_from_quantitative_analysis}
In Table \ref{tab:pcover}a, as $P_{cover}$ of training set increases ($\lambda$ increases), all three models show higher coverage for the input table records and generate longer sentences. Notably, although $R_{hallu}$ of training set remains the same for all $\lambda$, we see a small growth on $R_{hallu}$ of model outputs while increasing $\lambda$\footnote{We assume this is due to the information loss when filtering the records, as a poorly-informed decoder may behave like an open-ended language model \cite{maynez2020faithfulness}.}. 

In Table \ref{tab:pcover}b, when decreasing $N_{keep}$ (truncating more sentences), the hallucination ratio $R_{hallu}$ in model outputs are reduced as the same metric drops in the training set. However, while reducing hallucination, $P_{cover}$ in training set and model outputs declines at the same time, which reminds us the potential trade-off in truncation.

We have applied dataset modification, i.e. filtering and truncation, on the training set and seen there might be a trade-off between reducing hallucination (decreasing $R_{hallu}$) and increasing table coverage (increasing $P_{cover}$). For faithful description generation, we aim to find a way to reduce $R_{hallu}$ in the training phase while keeping table coverage (almost) unchanged. We come up with pseudo training data augmentation in plan-based generation and instance selection based on faithfulness ranking according to the quantitative analysis in this section.

\section{Entity-centric Faithful Generation}
\label{model}
In this section, based on the alignment between source tables and reference text, we propose two methods for faithful generation: augmented plan based generation and training instance selection according to faithfulness ranking.

\subsection{Augmented Plan-based Generation}
\label{sec:augmented_plan}
\subsubsection{Two-step Plan-based Generation}
\citet{moryossef2019step} have shown a \emph{two-step generator} with a separate text planner could improve generation faithfulness on WebNLG, a human curated dataset where the reference text barely contains any hallucination. The number of input table records in WebNLG is always smaller than 8. They view the input records as directed connected graph and treat depth first search (DFS) trajectory as the intermediate plan. 
However, in real world open domain setting, the input tables might contain diverse distinct table records which makes DFS trajectory plans very time-consuming to train. Moreover, the reference text in the training data might contain hallucinations which can not be aligned with any record in the table, which greatly hinder the generation fidelity.

To this end, we firstly extend the usage of plan-based method to the open domain setting using serialized plans.
We decompose the generation into two stages: table-to-plan and plan-to-text by introducing a trainable text planer.
As exemplified in Fig \ref{fig:plan_based_generation}, for the text planner,
we propose a neural plan generator (`$\mathrm{R}$'$\rightarrow$`$\mathrm{R}$+$\mathrm{P}$') 
which transforms the input table records to the serialized plans. The plans contain sentence planning (`SEP' token) and order planning of table records in each sentence. 
Specifically we constrain the plan generator to output only the attribute names that appear in the input table besides the `SEP' token and then add the paired attribute value for each generated attribute name by post editing
\footnote{In testing, we also post-process the learned plans to avoid empty sentence plans (i.e. two consecutive `SEP' tokens) and repetitive table records (i.e. limiting each table record to appear only once except `Name\_ID'). \label{plan-postprocess}}.
For the neural planner, we can easily use state-of-the-art sequence-to-sequence models thanks to textual serialized plans.

\subsubsection{Pseudo Parallel Training Corpus} 
\label{sec:pseudo_corpus}
For text realization, we acquire descriptions by feeding the plans to a plan-to-text generator.
In this process, we introduce augmented plans and pseudo parallel training corpus to reduce hallucinations in the generation.
We incorporate the auxiliary entity information extracted from reference text into the augmented plans (`Aug-plan', `$\mathrm{R}$+$\mathrm{E}$+$\mathrm{P}$' in Fig \ref{fig:plan_based_generation}) and feed them to the plan-to-text generator in the training phase.
In this way, the augmented plans and related text form a pseudo parallel training corpus which does not contain any hallucinated entity ($R_{hallu}=0$ in this scenario). According to the experimental results, the pseudo parallel corpus greatly reduces the hallucinations in the generated descriptions, which is consistent with our findings in Sec \ref{sec:insights_from_quantitative_analysis}.

Please refer to Table \ref{tab:plan_results} for the different input forms of the plan-to-text generator in training and evaluation phases.

\subsubsection{Unsupervised generation}
As a byproduct of entity recognition, 
we also propose an unsupervised model which does not need parallel table-to-text corpus, instead it uses text-only resources and the corresponding recognized entities in text.
Concretely in training we only feed a sequence of entity mentions (`$\mathrm{E}$' without NER tags in Fig \ref{fig:plan_based_generation}) extracted from text to the generator. While testing we use the values of table records (`$\mathrm{R}$' without attribute names in Fig \ref{fig:plan_based_generation}) as input.

\begin{table}[t]
\small
\setlength{\tabcolsep}{1.55mm}{
\begin{tabular}{l|ccccc}
\hline
Setting & BLEU & PARENT & $P_{cover}$ & $R_{hallu}(\downarrow)$ & LEN \\ \hline
R-1\% & \textbf{11.15} & \textbf{51.87} & 88.67\% & 0.087 & 50.1  \\
S-1\% & 7.40  & 49.66 & \textbf{90.27\%} & \textbf{0.015} & 32.1  \\ \hline
R-5\% & \textbf{26.76} & 52.46 & 98.44\% & 0.094 & 61.7 \\ 
S-5\% & 11.80 & \textbf{59.25} & \textbf{98.86\%} & \textbf{0.018} & 46.7 \\ \hline
R-10\% & \textbf{26.84} & 52.73 & 98.55\% & 0.096 & 62.2 \\ 
S-10\% & 12.52 & \textbf{59.63} & \textbf{99.17\%} & \textbf{0.020} & 56.7 \\
\hline
\end{tabular}}
\caption{Model performances of different downsampling methods (Sec \ref{sec:instance-select}). `R', `S' means randomly or ranking-based instance selection methods respectively.} 
\label{tab:fewshot}
\end{table}

\begin{table*}[t!]
\small
\centering
\setlength{\tabcolsep}{2mm}{
\begin{tabular}{l|ccccccccc}
\hline
ID & Setting & Comp. & Train Input & Test\&Dev Input & BLEU & PARENT & $P_{cover}$(\%,$\uparrow$) & $R_{hallu}$($\downarrow$) & LEN\\\hline
1& Baseline (E2E) & -- & $\mathrm{R}$ & $\mathrm{R}$  & 31.16 & 52.54 & 98.31 & 0.099 & 81.6\\\hline
2& Unsupervised & Yes & $\mathrm{E}$ & $\mathrm{R}$ & \textbf{17.28} & \textbf{43.04} & \underline{\textbf{97.91}} & \underline{\textbf{0.029}} & 50.0\\
3& Unsupervised & No & $\mathrm{E}$ & $\mathrm{E}$ & \underline{48.27} & \underline{63.94} & -- & -- & 82.4\\\hline
4& Plan & Yes & $\mathrm{R}$+$\mathrm{P}_{gold}$ & $\mathrm{R}$ & \textbf{18.23} & 51.53 & 99.48 & 0.069 & 46.1 \\
5& Plan & Yes & $\mathrm{R}$+$\mathrm{P}_{gold}$ & $\mathrm{R}$+$\mathrm{P}_{learn}$ & 7.67 & \underline{\textbf{54.81}} & \underline{\textbf{99.66}} & \underline{\textbf{0.053}} & 53.2\\
6& Plan & No & $\mathrm{R}$+$\mathrm{P}_{gold}$ & $\mathrm{R}$+$\mathrm{P}_{gold}$ & \underline{34.86} & 54.25 & 97.76 & 0.097 & 81.6\\\hline
7& Aug-plan & Yes & $\mathrm{R}$+$\mathrm{E}$+$\mathrm{P}_{gold}$ & $\mathrm{R}$ & 12.10 & 50.70 & 98.68 & 0.015 & 35.8\\
8& Aug-plan & Yes& $\mathrm{R}$+$\mathrm{E}$+$\mathrm{P}_{gold}$ & $\mathrm{R}$+$\mathrm{P}_{learn}$ & \textbf{17.12} & \textbf{56.75} & \underline{\textbf{99.73}} & \underline{\textbf{0.006}} & 39.7\\
9& Aug-plan & No & $\mathrm{R}$+$\mathrm{E}$+$\mathrm{P}_{gold}$ & $\mathrm{R}$+$\mathrm{P}_{gold}$ & 29.12 & 57.84 & 97.98 & 0.029 & 60.6\\
10& Aug-plan & No & $\mathrm{R}$+$\mathrm{E}$+$\mathrm{P}_{gold}$ & $\mathrm{R}$+$\mathrm{E}$+$\mathrm{P}_{gold}$ & \underline{53.33} & \underline{66.12} & -- & -- & 81.6\\\hline
\end{tabular}}
\\Human Evaluation (Sec \ref{sec:human_evaluation})- Ratio of hallucinated words: 
\\ End-to-end [ID=1]: 20.9\%; Unsupervised [ID=2]: 12.0\%; Plan-based [ID=5]: 15.3\%; Augplan-based [ID=8]: 8.1\% 
\caption{The performances of BART-based unsupervised and plan-based models on \textsc{Wikiperson}. The baseline is an end-to-end (E2E) trained BART model in Table \ref{tab:benchmark}. `$\mathrm{R}$', `$\mathrm{E}$', `$\mathrm{P}_{gold}$' and `$\mathrm{P}_{learn}$' represent table records, recognized entities, gold plans extracted from reference text and learned plans respectively as shown in Fig \ref{fig:plan_based_generation}. `Comp.' indicates the comparability to baseline as gold plan $\mathrm{P}_{gold}$ or entities $\mathrm{E}$ are actually not accessible while testing. In each setting, we underline the best scores, and show the best scores comparable to baseline in bold.}
\label{tab:plan_results}

\end{table*}

\begin{table}[t!]
    \centering
    \small
    \begin{tabular}{lcc}
\hline
Model & BLEU & PARENT \\\hline
\multicolumn{3}{c}{RNN Structure}\\
RNN \cite{BahdanauCB14} & 22.24 & 43.41 \\
Structure-aware \cite{liu2017table} & 22.76 & 46.47 \\
PG-Net \cite{see2017get} & \textbf{23.56} & 50.14 \\
KB Description \cite{wang2018describing} & 16.20 & \textbf{51.03}  \\\hline
\multicolumn{3}{c}{Transformer Structure}\\
Transformer \cite{wang2020towards} & 24.57 & 51.87 \\
+Content Matching \cite{wang2020towards} & 24.56 & 53.06 \\
Transformer (Our Implementation) & \textbf{24.63} & 51.86 \\
w/ Plan & 7.15 & 52.92 \\
w/ Augmented Plan &  14.56 & \textbf{54.78} \\\hdashline
BART (Our Implementation) & \underline{\textbf{31.16}} & 52.54 \\
w/ Plan & 7.67 & 54.81 \\
w/ Augmented Plan & 17.12 & \underline{\textbf{56.75}} \\\hline
    \end{tabular}
    \caption{The automatic evaluation of augmented plan-based method and other baselines on \textsc{Wikiperson} dataset. Note that in this dataset, the reference text may be noisy due to the hallucinated content, so BLEU scores can not measure the generation fidelity. The proposed augmented plan-based generation enhances faithfulness in both Transformer and BART structures and achieve the best PARENT scores.}
    \label{tab:plan_vs_stoa}
\end{table}

\subsection{Instance Selection for Efficient Training}
\label{sec:instance-select}
We have shown in the last sections that the divergence between source tables and reference text is one of the major reasons that hallucination exist in the generated text. We thus wonder if given high quality training sets, how much performance gain we could expect from the end-to-end trained models.
We focus on reducing hallucination, which is shown to be the bottleneck for faithful generation. We rank the training instances with instance-level hallucination ratio $r_{hallu}=n_{hallu}/l$ (lower is better)
\footnote{$n_{hallu}$ and $l$ are the number of hallucinated entities and reference length as introduced in Sec \ref{sec:entity_centric_evaluation}.}. 
Then we view the top-1\%/5\%/10\% ranking data as `selected' training sets (`S'), and compare them to randomly sampled training sets (`R') in Table \ref{tab:fewshot}.
`S-n\%' can be viewed as a training set which has higher quality than the corresponding `R-n\%'.
When n grows in S-n\%, we see the increase of table coverage $P_{cover}$, generation length and also a slight increase on $R_{hallu}$, which consistently outperform their counterparts R-n\%.
S-5\% even outperforms full-dataset baseline (ID=1 in Table \ref{tab:plan_results}) by 7 PARENT points.

\begin{table}[t!]
    \centering
    \small
    \begin{tabular}{lccccc}
\hline
Model & 100 & 250 & 500 & 2500 & All \\\hline
& \multicolumn{4}{c}{PARENT $(\uparrow)$}\\
BART & 25.48 & 41.62 & 47.34 & 51.72 & 52.54 \\
w/ Plan & 37.23 & 47.57 & 50.92 & 52.67 & 54.81 \\
w/ Aug-Plan & \textbf{40.46} & \textbf{50.32} & \textbf{52.38} & \textbf{54.13} & \textbf{56.75} \\\hline
& \multicolumn{4}{c}{$P_{cover} (\%) (\uparrow)$ }\\
BART & 77.46 & 80.32 & 82.48 & 89.02 & 98.31 \\
w/ Plan & \textbf{79.21} & 83.53 & \textbf{88.59} & 92.19 & 99.66 \\
w/ Aug-Plan & 78.93 & \textbf{84.05} & 88.39 & \textbf{92.74} & \textbf{99.73} \\\hline
& \multicolumn{4}{c}{$R_{hallu} (\downarrow)$}\\
BART & 0.077 & 0.080 & 0.085 & 0.089 & 0.099\\
w/ Plan & 0.043 & 0.045 & 0.048 & 0.048 & 0.053\\
w/ Aug-Plan & \textbf{0.008} & \textbf{0.007} & \textbf{0.005} & \textbf{0.005} & \textbf{0.006} \\\hline
    \end{tabular}
    \caption{Model performance of plan-based models in the few shot learning setting. 100/250/500/2500 represent the size of training instances.}
\end{table}

\section{Experiments}
\subsection{Datasets}
The \textsc{Wikiperson} dataset\footnote{\url{https://github.com/EagleW/Describing_a_Knowledge_Base}} \cite{wang2018describing}  contains 250186, 30487,
and 29982 \emph{(table, text)} pairs in training, dev and test sets respectively, which is exactly the same setting as \citet{wang2018describing} and \citet{wang2020towards}. The average sentence number in this dataset is 4.33 (88.3 tokens).

\subsection{Automatic and Human Evaluation}
\label{sec:human_evaluation}
Apart from the proposed metrics $P_{cover}$ and $R_{hallu}$, we also report the model performances on BLEU and PARENT. PARENT is hybrid measurement which not only encourages n-gram overlap with reference text but also rewards the generation with high coverage on the source table, thus it can be used to measure the generation fidelity.

For human evaluation, as we have mentioned in Sec \ref{sec:insights_from_quantitative_analysis}, the proposed two metrics have high precision in identifying hallucinations, however it may not find all hallucinations in the generation.
For four models in Table \ref{tab:plan_results}: baseline (ID=1), unsupervised (ID=2), plan-based (ID=5) and augplan-based (ID=8) models,
we conduct human evaluation\footnote{Here we omit human evaluations on the fluency and table record coverage as we have seen all the outputs have high coverage (Table \ref{tab:plan_results}) and good fluency (thanks to BART).}
by randomly sampling 100 generations from each model and hiring three annotators to independently annotate $any$ hallucinated word in generation. The Fleiss' kappa of human judgement is 0.52.
Through the human judgement, we can obtain the ratio of hallucinated words.
We then calculate the overall hallucinated word ratio by averaging the annotators' decisions.

\begin{figure*}[t!]
    \centering
    \includegraphics[width=1.0\linewidth]{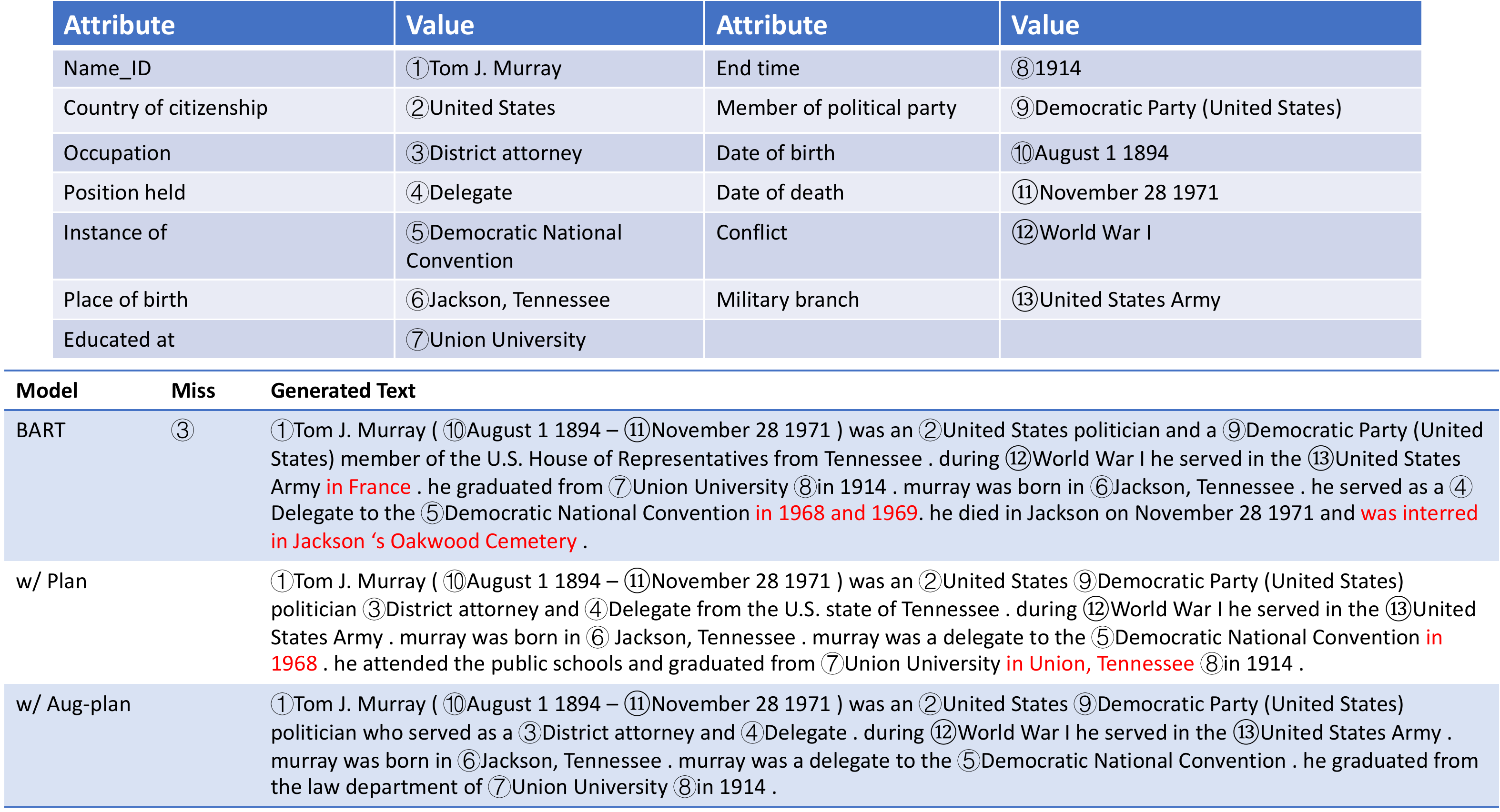}
    \caption{An example from the test set of \textsc{Wikiperson}. We mark the hallucinated content in red. Note that the marked content is \emph{factually} wrong, e.g. in end of `BART' generated text, Murray was actually interred in `the city's Hollywood Cemetery' rather than `Jackson's Oakwood Cemetery', which shows that those hallucinated content actually comes from the misguidance of hallucinations in the training stage, rather than the implicit knowledge acquisition of large-scale pre-trained models.   }
\label{fig:case_study}
\end{figure*}

\subsection{Experimental Settings}
In this paper we implement Pointer-Generator (PG-Net) \cite{see2017get}, Transformer \cite{vaswani2017attention} and BART \cite{lewis2019bart} models on \textsc{Wikiperson} dataset.
Our implementation for Transformer and BART is based on fairseq \cite{ott2019fairseq} \footnote{\url{https://github.com/pytorch/fairseq}}.
For PG-Net, our implementation is based on a pytorch version re-implementation on github\footnote{\url{https://github.com/atulkum/pointer_summarizer}}.
For Transformer model, we use the same parameter setting as \citet{wang2020towards} (with copy mechanism). The hidden units of the multi-head attention and the feed-forward layer are set to 2048. The embedding size is set to 512. The number of heads is set to 8, and the number of Transformer blocks is 3. Beam size is set to be 5. For other parameters except learning rate, we use the default setting in fairseq according to \citet{vaswani2017attention}.
For BART, we use the pretrained BART (large) architecture in fairseq which has 12 layers.
For PG-Net, we set the embedding size, hidden size and beam size as 256, 512 and 5 respectively.
For all three models, we use adam optimizer \cite{kingma2014adam} ($\beta_1=0.9, \beta_2=0.999$). The learning rates are set as 1e-5, 1e-5, 3e-5 for PG-Net, Transformer and BART models respectively.
The vocabulary size is limited to 50k for every model.

\section{Analyses and Discussions}
\subsection{Augmented Plan-based Method}
\subsubsection{Comparsion with other baselines}
We compare the proposed augmented plan-based model with the other baselines in Table \ref{tab:plan_vs_stoa}.
The plan-based method outperform the state-of-the-art baseline \cite{wang2020towards}. Note that our implemented transformer has similar model capacity with \citet{wang2020towards}, i.e. the same hidden dimension setting.
The experiments with BART model show that the proposed method also works well with large-scale pre-trained models.

\subsubsection{A closer look at plan-based models}
We list the model input in the plan-to-text stage (Sec \ref{sec:pseudo_corpus}) of plan-based generation while training or testing the model in Table \ref{tab:plan_results}.
We summarize our findings from the automatic and human evaluations in Table \ref{tab:plan_results} as follows:

\begin{figure}[t!]
    \centering
    \includegraphics[width=1.0\linewidth]{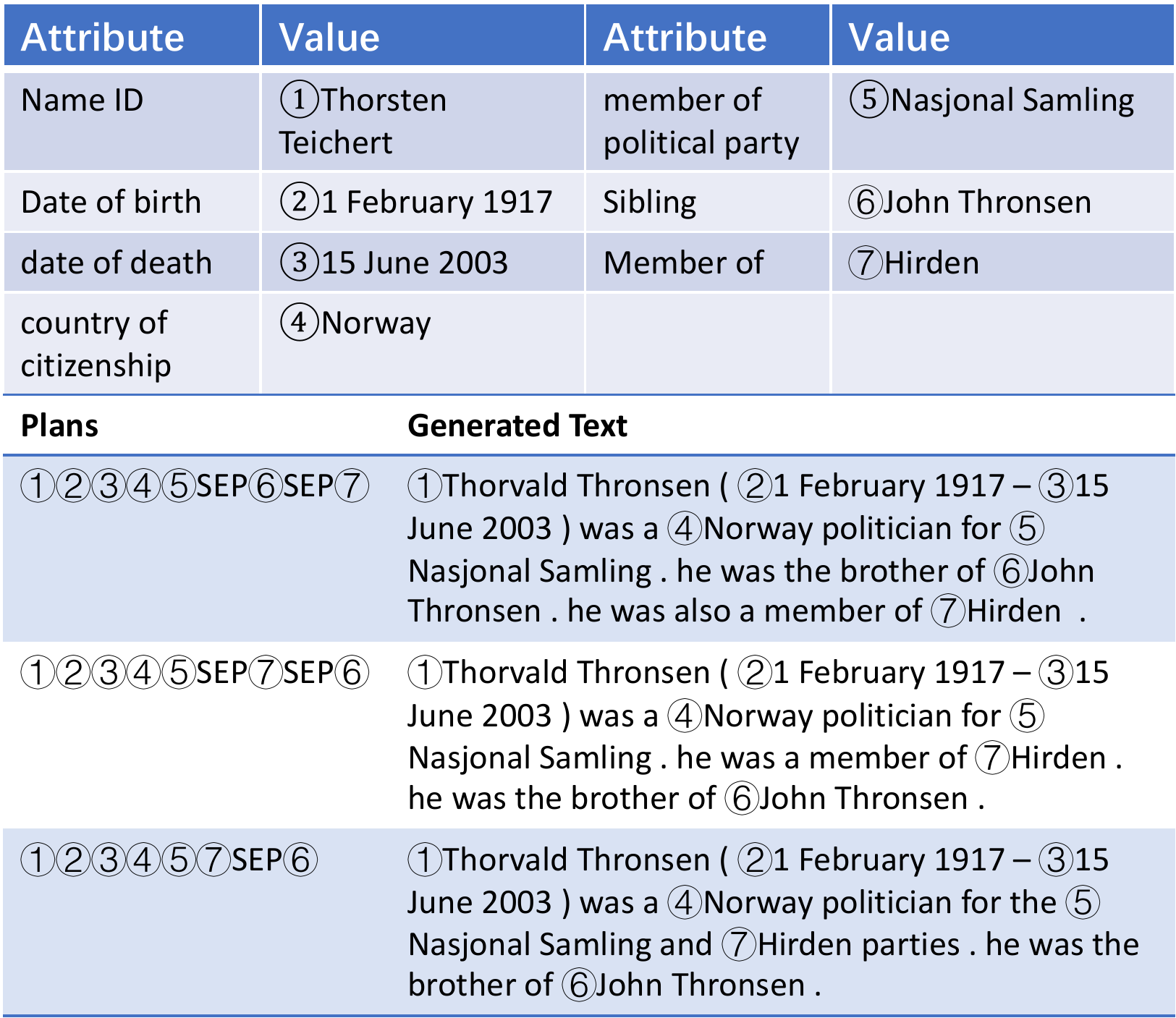}
    \caption{A demonstration of different model outputs given different plans. Here we show the generated text from augmented plan-based models on the test set. The generated text perfectly follows the given plans.}
\label{fig:plan}
\end{figure}

\noindent 1) Both augmented plan-based and unsupervised models greatly reduce generation hallucination as we create pseudo parallel corpus by explicitly adding hallucinated entities in reference text while training. The new model inputs perfectly match the references as all the entities in reference text appear in the new inputs ($R_{hallu}=0$ in training set). This phenomenon comports with our findings in Table \ref{tab:pcover} (b): decreasing $R_{hallu}$ in training set correspondingly reduce hallucination in the evaluation phase. 

\noindent 2) For testing with gold plans (ID=6,9,10), we use the gold plans extracted from test reference text \emph{without} the post-processing mentioned in Footnoot \ref{plan-postprocess}\footnote{This means these gold plans may contain two consecutive `SEP' tokens and repetitive table records.}, in order to get better BLEU scores, however at the same time sacrificing fidelity ($R_{hallu}$ in ID=5,6 and ID=8,9). 
This suggests BLEU may not be a good metric for faithfulness evaluation in the loosely aligned table-to-text datasets\footnote{PARENT is a better option which encourages the overlap with source table. However it also rewards the ngram overlap with reference text, which may contain hallucination. We would suggest human evaluation for faithfulness analysis.}.

\noindent 3) In ID = 4 or 7, plan-based models would generate one-sentence output given the original table records `$\mathrm{R}$', which means that the generation sticks to its plan\footnote{No `SEP' token in the input plan means we want to describe each record in the input sequentially in one sentence.}.
We find the same tendency in case studies, showing model's potentials to allow user control. 
Furthermore, we find the explicit planing would also reduce the repetitions of table records.

\subsubsection{Plan-based Generation helps few shot learning}
In Table \ref{tab:fewshot}, we show that plan-based generation can make the generated text more faithful as well as have higher coverage compared with end-to-end training. Augmented plans further reduce the hallucinations in the generated text (much smaller $R_{hallu}$) compared with its variant without auxiliary entities.

\subsection{Case Studies}
Firstly, we show an example taken from the test set of \textsc{Wikiperson} dataset in Fig \ref{fig:case_study}. The baseline model, BART trained in the end-to-end fashion misses the `occupation' attribute in the source table, while the other two models cover all the attributes in the table.
The augmented plan-based generation does not contain any hallucinated content while some factually wrong information exist in the baseline and plan-based methods.

Then we show in Fig \ref{fig:plan} that the generated text from augmented plan-based method strictly sticks to the corresponding plans. It also shows that the augmented plan-based method has the potential to generate more diverse content by exploiting different plans. 

\section{Related Work}

\noindent \textbf{Faithfulness Evaluation}: Prior work evaluates the faithfulness in generation by human \cite{chen2018fast} or automatic metrics using natural language inference \cite{kryscinski2019evaluating,falke2019ranking}, information extraction \cite{zhang2019optimizing,goodrich2019assessing} or question answering \cite{scialom2019answers,eyal2019question,wang2020asking}. For faithfulness evaluation, prior work introduces soft constraints \cite{tian2019sticking,wang2020towards}, template-like \cite{liu2017large,DBLP:conf/emnlp/WisemanSR18,shen2020neural,li2020posterior,ye2020variational} or controllable generation \cite{peng2018towards,fan2019strategies,shen2019select,parikh2020totto}, which encourage the overlap between given structured data and generated text. 
Some work also incorporate source input (table or text) in evaluation\cite{liu-etal-2019-towards-comprehensive,dhingra2019handling,wang-etal-2020-anchor}.

\noindent \textbf{Plan-base Generation}: Most work before deep learning era treats data-to-text as two sub-tasks: content selection and surface realization \cite{reiter1997building,duboue2002content,barzilay2005collective,barzilay2006aggregation,belz2008automatic}, which carefully learns the alignment between data and text\cite{liang2009learning,angeli2010simple,kim2010generative,konstas2013global}. Recently, end-to-end learning becomes a trend \cite{MeiBW16,chisholm2017learning, DBLP:conf/naacl/KaffeeEVGLHS18,jhamtani2018learning,bao2018table,liu2019hierarchical,duvsek2020evaluating} in this field. Among them, some work introduces differentiable planning modules \cite{DBLP:conf/aaai/ShaMLPLCS18,DBLP:journals/corr/abs-1810-02889,puduppully2018data}.
Our paper focuses on a two-stage generation which incorporate separate text planner \cite{ferreira2019neural,moryossef2019improving,ma-etal-2019-key}. 
The separate planning methods could be easier to deploy and debug in the real world scenarios, it also shares the burden of end-to-end learning with two separate modules.

\section{Conclusion}
\label{conclusion}
We try to evaluate faithfulness in table-to-text generation by entity-centric statistics on table coverage and hallucination in text, which help us analyze the correlation between training data quality and generation fidelity. Then we accordingly propose two methods based on the analysis: augmented training by incorporating auxiliary entity information and instance selection based on faithfulness ranking. We show these methods improve generation faithfulness on both automatic and human evaluations.

\section*{Acknowledgments}
We would like to thank the anonymous reviewers for the helpful discussions and suggestions.  
Our work is supported by the National Key Research and Development Program of China No.2020AAA0106701, NSFC project No.U19A2065, No.61876004, and Beijing Academy of Artificial Intelligence (BAAI).

\small
\bibliography{aaai21}

\end{document}


\maketitle

\begin{figure*}
    \centering
    \includegraphics[width=0.8\linewidth]{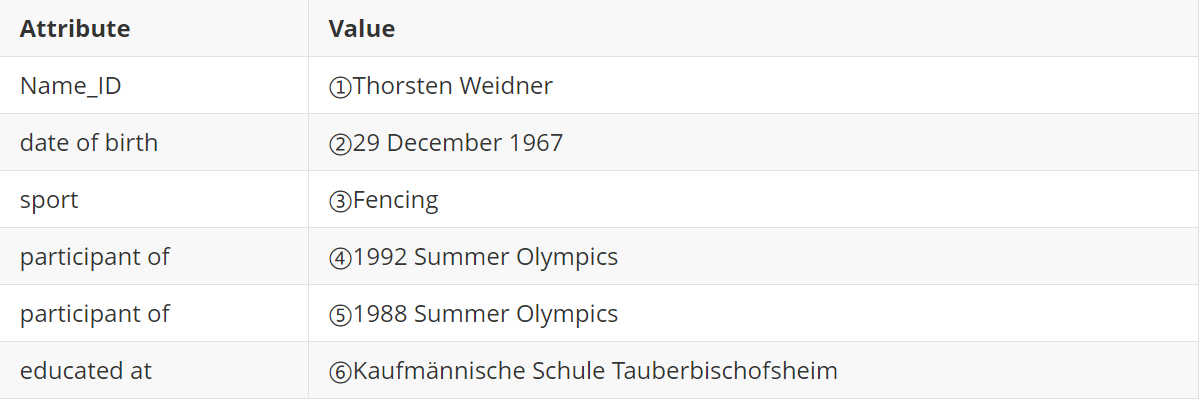}
    \includegraphics[width=0.8\linewidth]{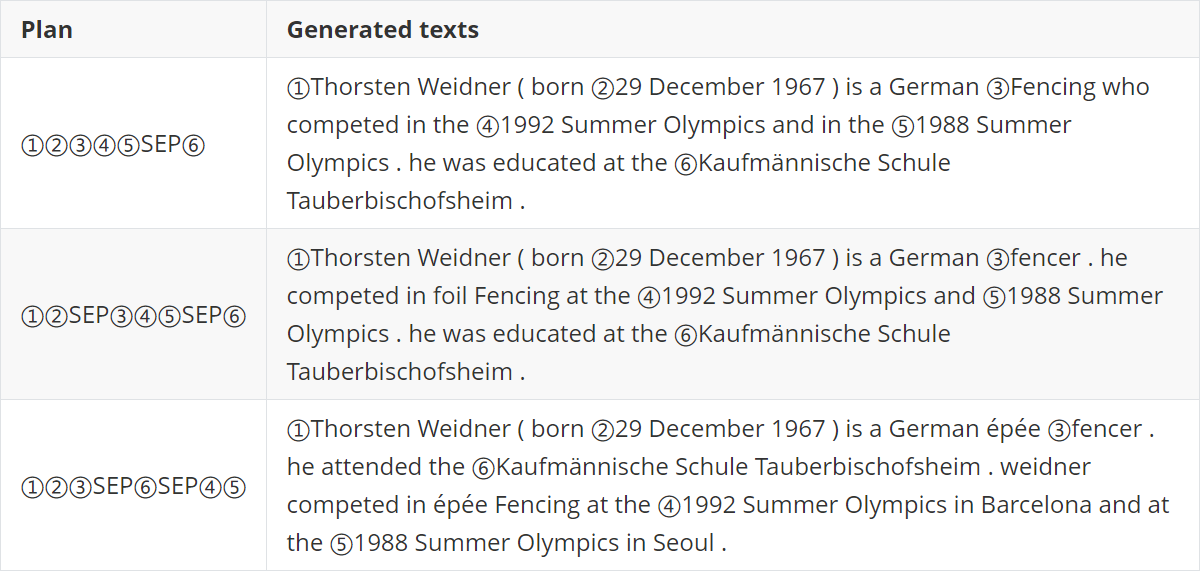}
    \caption{The case study of user control in plan-based generation. We show the diverse model outputs from the proposed augplan-based model, according to designated plans.}
    \label{fig:plan1}
\end{figure*}

\begin{figure*}
    \centering
    \includegraphics[width=0.8\linewidth]{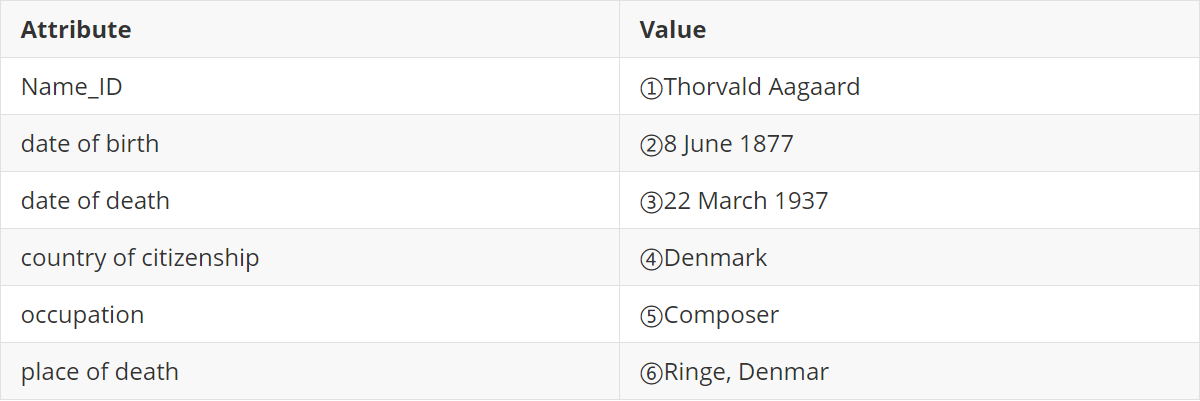}
    \includegraphics[width=0.8\linewidth]{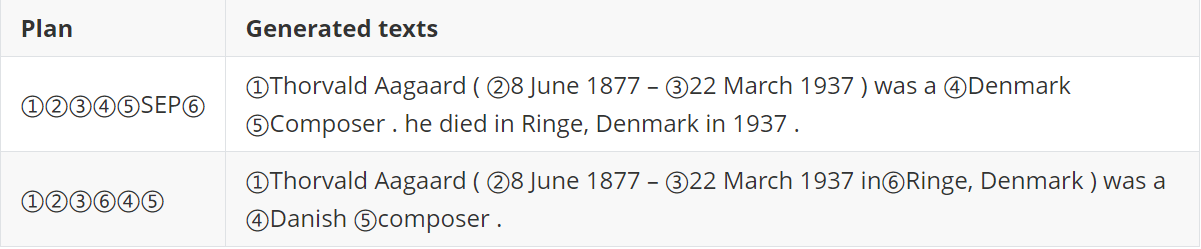}
    \caption{The case study of user control in plan-based generation. Same setting as Fig \ref{fig:plan1}.}
    \label{fig:plan2}
\end{figure*}

\begin{figure*}
    \centering
    \includegraphics[width=0.8\linewidth]{pics/plans/plan.1.table.png}
    \includegraphics[width=0.8\linewidth]{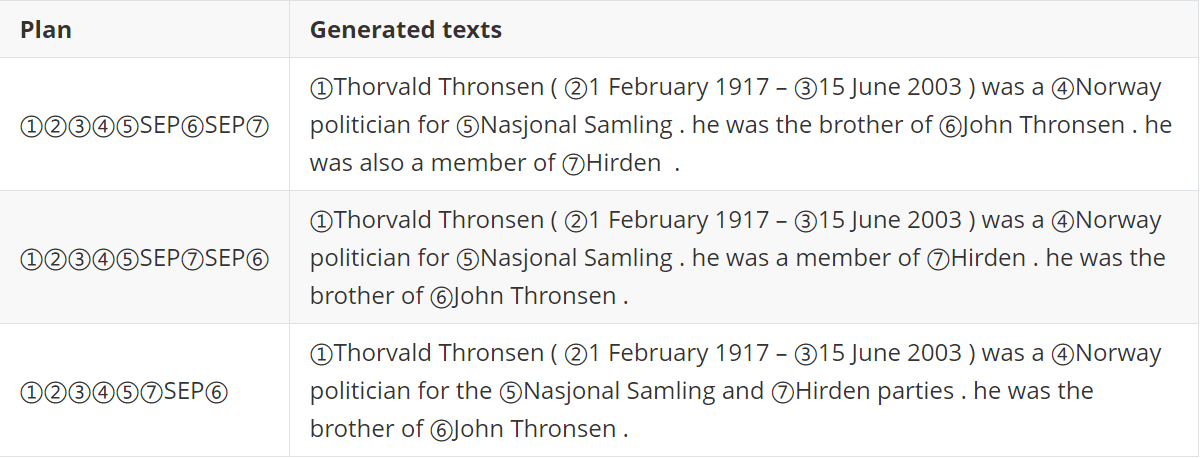}
    \caption{The case study of user control in plan-based generation. Same setting as Fig \ref{fig:plan1}.}
    \label{fig:plan3}
\end{figure*}

\begin{figure*}
    \centering
    \includegraphics[width=0.8\linewidth]{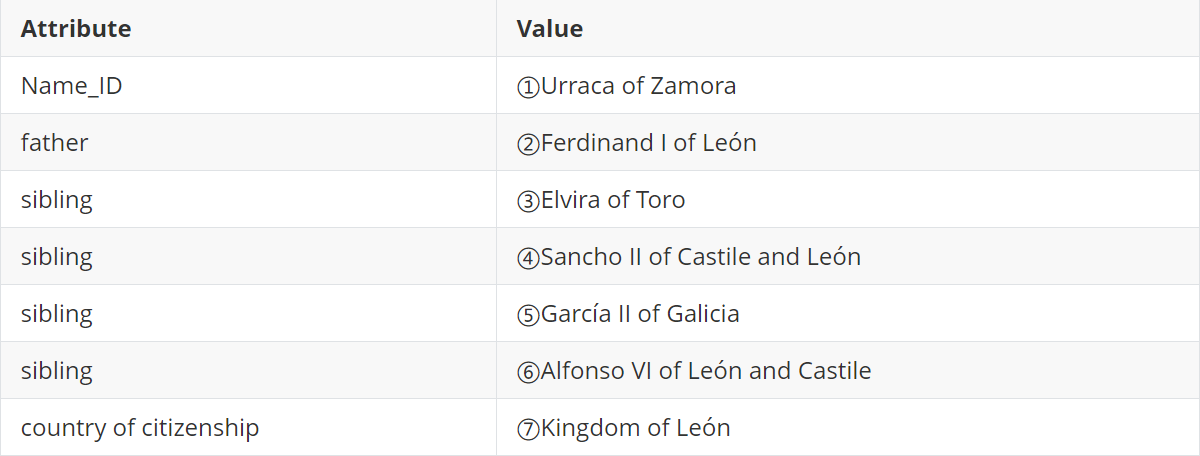}
    \caption{The source table about `Urraca of Zamora'.}
    \label{fig:model_table1}
\end{figure*}

\begin{figure*}
    \centering
    \includegraphics[width=0.8\linewidth]{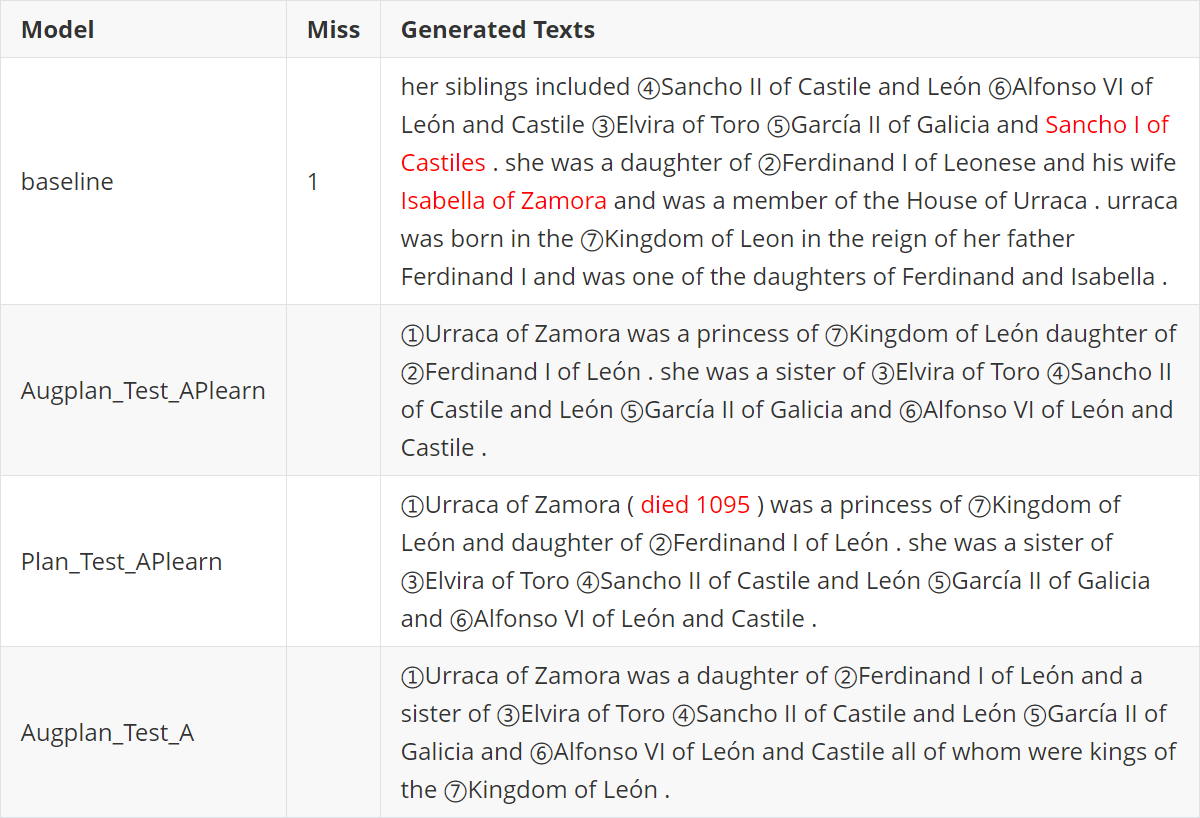}
    \includegraphics[width=0.8\linewidth]{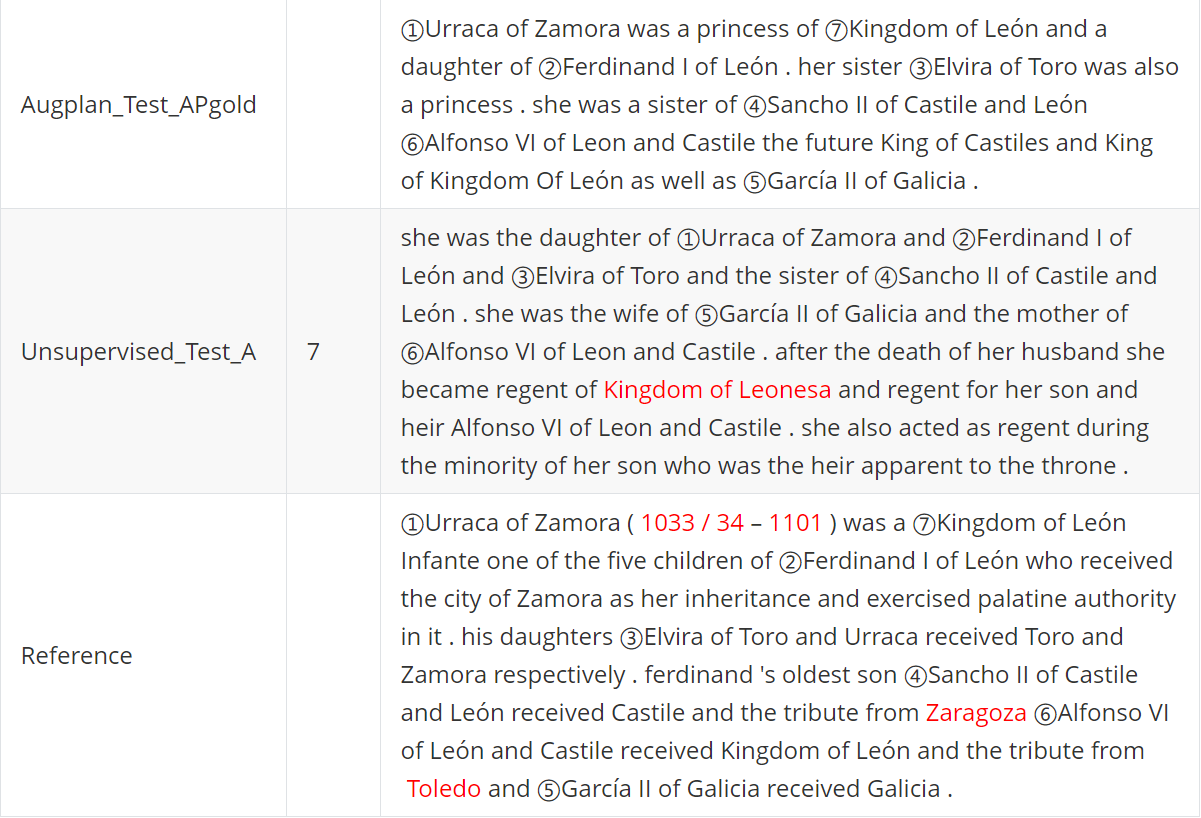}
    \caption{The model outputs and reference text for source table about `Urraca of Zamora' in Fig \ref{fig:model_table1}. We mark the hallucinated words in red and also highlight the missing table records.}
    \label{fig:model_output1}
\end{figure*}

\begin{figure*}
    \centering
    \includegraphics[width=0.8\linewidth]{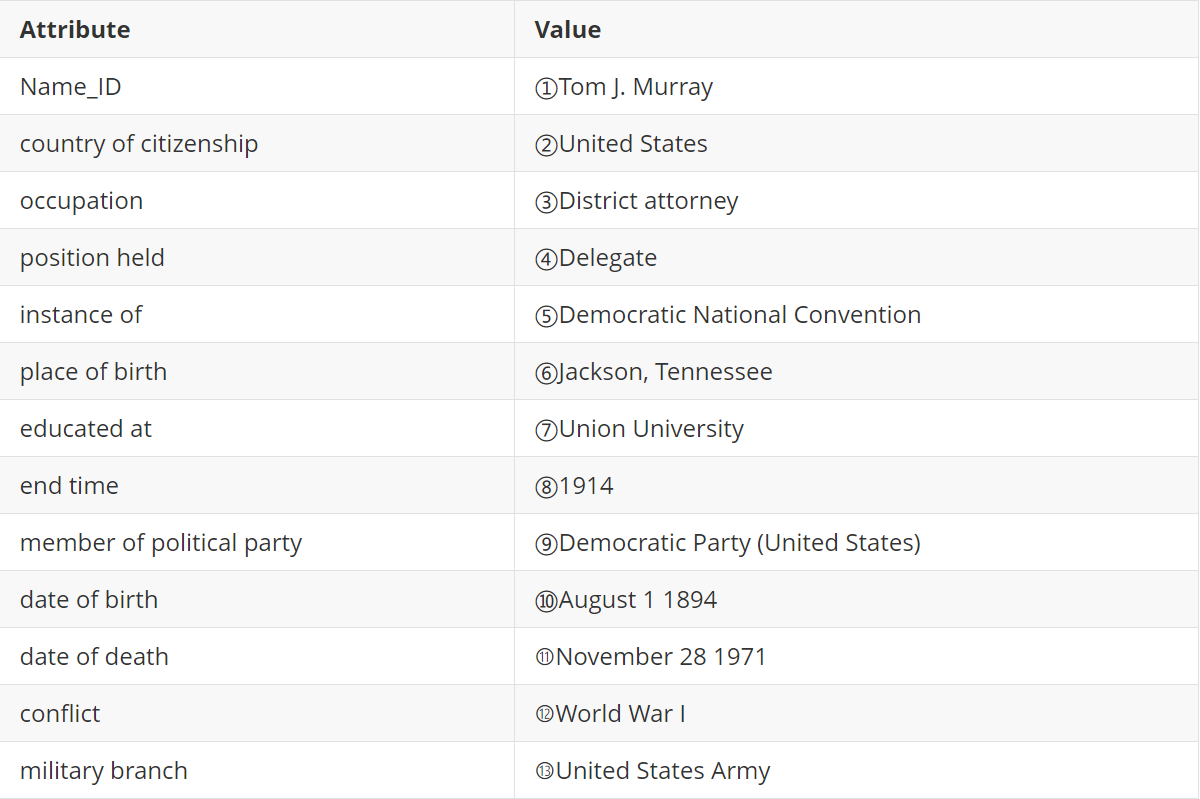}
    \caption{The source table about `Tom J. Murray'.}
    \label{fig:model_table2}
\end{figure*}

\begin{figure*}
    \centering
    \includegraphics[width=0.8\linewidth]{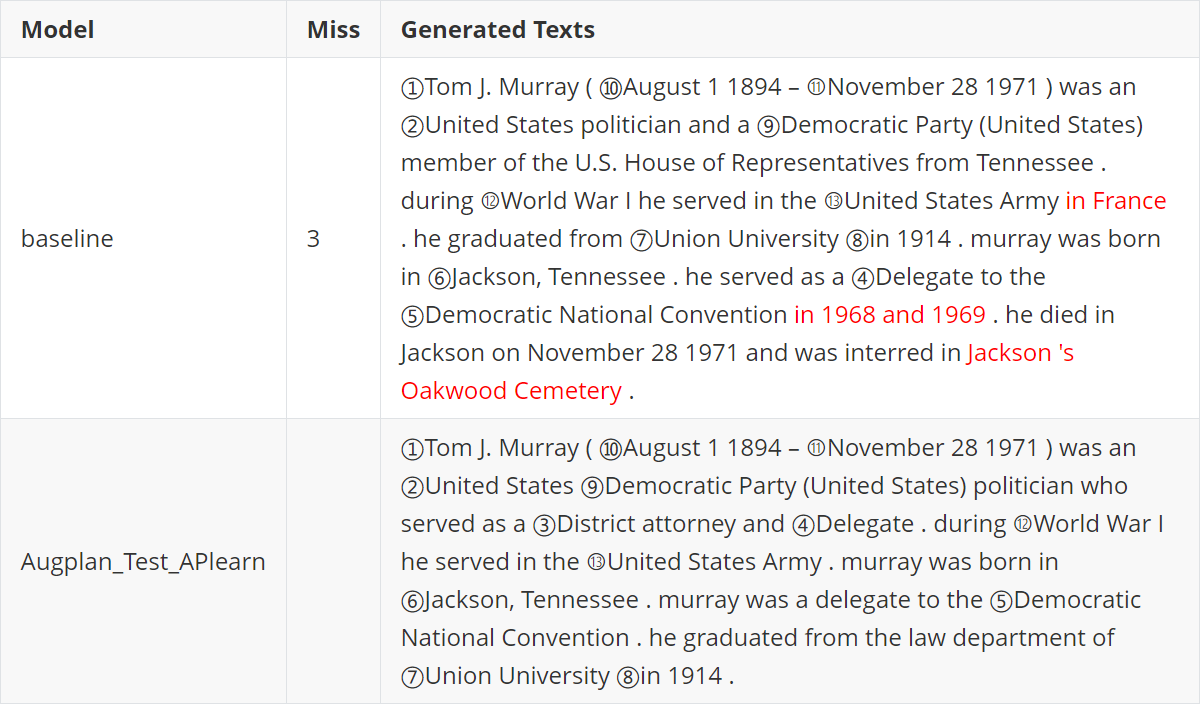}
    \includegraphics[width=0.8\linewidth]{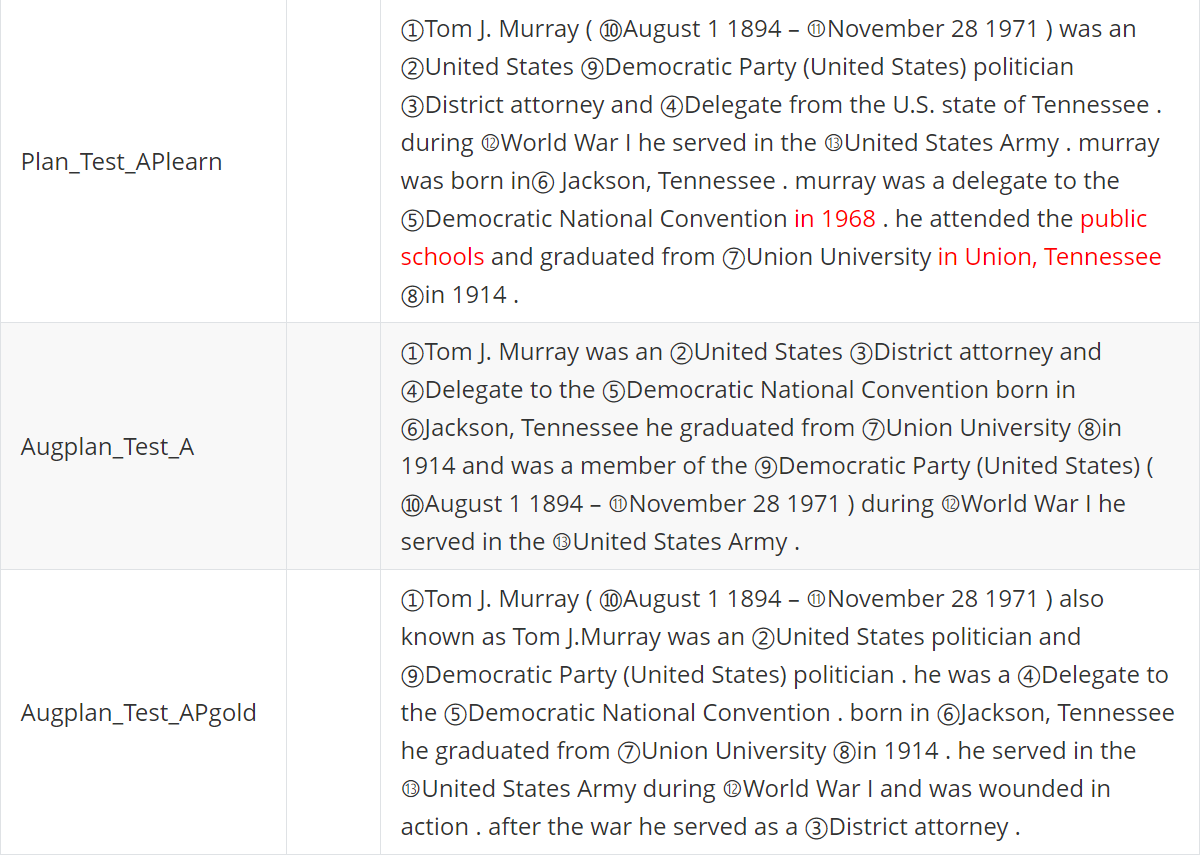}
    \includegraphics[width=0.8\linewidth]{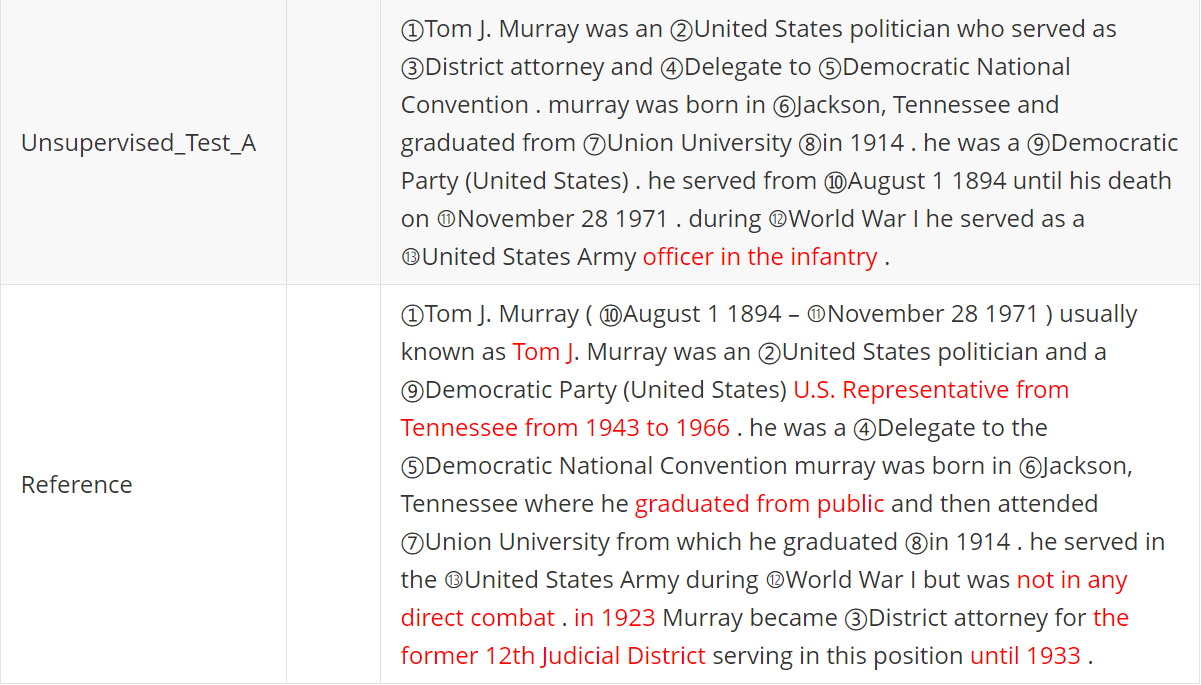}
    \caption{The model outputs and reference text for source table about `Tom J. Murray' in Fig \ref{fig:model_table2}.}
    \label{fig:model_output2}
\end{figure*}

\begin{figure*}
    \centering
    \includegraphics[width=0.8\linewidth]{pics/cases/case.3.table.png}
    \caption{The source table about `Tim Sherwood'.}
    \label{fig:model_table3}
\end{figure*}

\begin{figure*}
    \centering
    \includegraphics[width=0.8\linewidth]{pics/cases/case.3.text1.png}
    \includegraphics[width=0.8\linewidth]{pics/cases/case.3.text2.png}
    \includegraphics[width=0.8\linewidth]{pics/cases/case.3.text3.png}
    \includegraphics[width=0.8\linewidth]{pics/cases/case.3.text4.png}
    \caption{The model outputs and reference text for source table about `Tim Sherwood' in Fig \ref{fig:model_table3}.}
    \label{fig:model_output3}
\end{figure*}

\bibliography{aaai21}
\bibliographystyle{aaai21}